\documentclass[10pt,twocolumn]{Myarticle}

\usepackage{iccv}
\usepackage{times}
\usepackage{epsfig}
\usepackage{microtype}
\usepackage{amsmath}
\usepackage{amssymb , bbm}
\usepackage[inline]{enumitem}
\usepackage{epstopdf}
\usepackage{caption}
\usepackage{subcaption}
\usepackage[font={small}]{caption}
\usepackage{tabularx}
\usepackage{booktabs}
\usepackage{multirow}
\usepackage{pifont}

\iccvfinalcopy
\ificcvfinal\fi

\usepackage{enumitem}
\setitemize[0]{leftmargin=10pt}

\newenvironment{tight_itemize}{
\begin{itemize}[leftmargin=10pt]
  \setlength{\topsep}{0pt}
  \setlength{\itemsep}{2pt}
  \setlength{\parskip}{0pt}
  \setlength{\parsep}{0pt}
}{\end{itemize}}

\title{Universal Semi-Supervised Semantic Segmentation}

\author{ Tarun Kalluri$^1$ \qquad Girish Varma$^1$ \qquad Manmohan Chandraker$^2$ \qquad C V Jawahar$^1$\\[0.5em]
$^1$Center for Visual Information Technology, IIIT Hyderabad \\ $^2$University of California, San Diego \\
{\tt\small tarun.05.kalluri@gmail.com}
}

\usepackage[pagebackref=true,breaklinks=true,letterpaper=true,colorlinks,bookmarks=false]{hyperref}

\newcommand{\figref}[1]{Figure~\ref{#1}}
\newcommand{\secref}[1]{Section~\ref{#1}}
\newcommand{\tabref}[1]{Table~\ref{#1}}
\renewcommand{\eqref}[1]{Eq~(\ref{#1})}
\newcommand{\LY}{\mathcal{Y}}
\newcommand{\LS}{\mathcal{L}}
\newcommand{\D}{\mathcal{D}}

\newcommand{\x}{\mathrm{x}}
\newcommand{\X}{\mathrm{X}}
\newcommand{\y}{\mathrm{y}}
\newcommand{\Y}{\mathrm{Y}}

\newcommand{\tick}{{\Large{{\ding{51}}}}}
\newcommand{\cross}{{\Large{{\ding{55}}}}}

\begin{document}

\maketitle
\thispagestyle{empty}

\begin{abstract}
In recent years, the need for semantic segmentation has arisen across several different applications and environments. However, the expense and redundancy of annotation often limits the quantity of labels available for training in any domain, while deployment is easier if a single model works well across domains. In this paper, we pose the novel problem of universal semi-supervised semantic segmentation and propose a solution framework, to meet the dual needs of lower annotation and deployment costs. In contrast to counterpoints such as fine tuning, joint training or unsupervised domain adaptation, universal semi-supervised segmentation ensures that across all domains: (i) a single model is deployed, (ii) unlabeled data is used, (iii) performance is improved, (iv) only a few labels are needed and (v) label spaces may differ. To address this, we minimize supervised as well as within and cross-domain unsupervised losses, introducing a novel feature alignment objective based on pixel-aware entropy regularization for the latter. We demonstrate quantitative advantages over other approaches on several combinations of segmentation datasets across different geographies (Germany, England, India) and environments (outdoors, indoors), as well as qualitative insights on the aligned representations\footnote{Code available at \href{https://github.com/tarun005/USSS\_ICCV19}{https://github.com/tarun005/USSS\_ICCV19}}.

\end{abstract}

\section{Introduction}
\label{sec:intro}

Semantic segmentation is the task of pixel level classification of an image into a predefined set of categories. State-of-the-art semantic segmentation architectures~\cite{long2015fully,badrinarayanan2015segnet,chen2018deeplab} pre-train deep networks for a classification task on datasets like ImageNet~\cite{deng2009imagenet, russakovsky2015imagenet} and then fine-tune on finely annotated labeled examples~\cite{cordts2016cityscapes,anue}. The availability of such large-scale labeled datasets has been crucial to achieve high accuracies for semantic segmentation in applications ranging from natural scene understanding~\cite{GARCIAGARCIA201841} to medical imaging~\cite{ronneberger2015u}. However, performance often suffers even in the presence of a minor domain shift. For example, a segmentation model trained on a driving dataset from a specific geographic location may not generalize to a new city due to differences in weather, lighting or traffic density. Further, a segmentation model trained on traffic scenes for outdoor navigation may not be applicable for an indoor robot.

\begin{figure}[!t]
\centering
  \includegraphics[width=\linewidth]{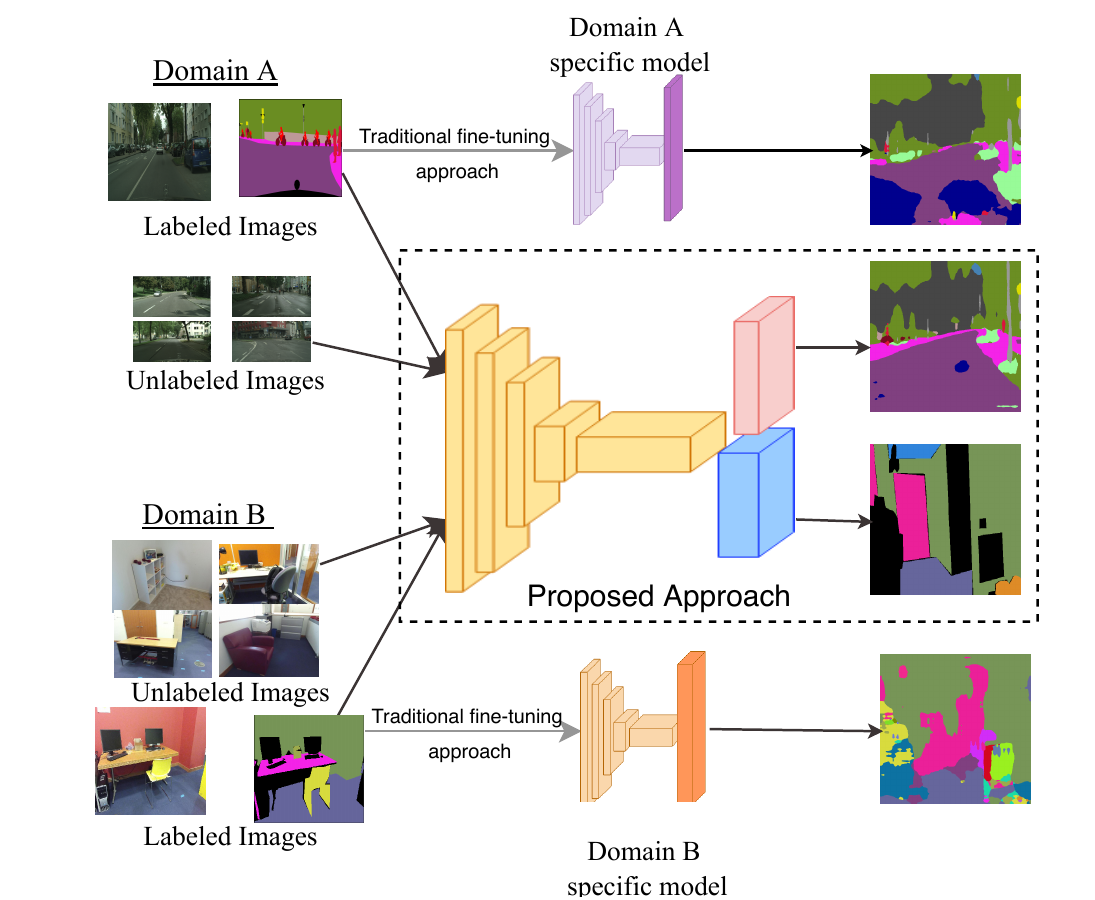}
\vspace{-2em}
    \caption{Proposed universal segmentation model can be jointly trained across datasets with different label spaces, making use of the large amounts of unlabeled data available. Traditional transfer learning based approaches typically require training separate models for each domain.}
\label{fig:intro_pic}
\vspace{-1.5em}
\end{figure}

While such domain shift is a challenge for any machine learning problem, it is particularly exacerbated for segmentation where human annotation is highly prohibitive and redundant for different locations and tasks. Thus, there is a growing interest towards learning segmentation representations that may be shared across domains. A prominent line of work addresses this through unsupervised domain adaptation from a labeled source to an unlabeled target domain \cite{hoffman2016fcns,tsai2018learning,chen2017no,murez2018image,carlucci2017auto}. But there remain limitations. For instance, unsupervised domain adaptation usually does not leverage target domain data to improve source performance. Further, it is designed for the restrictive setting of large-scale labeled source domain and unlabeled target domain. While some applications such as self-driving have large-scale annotated datasets for particular source domains (for example synthetic datasets like Synthia~\cite{RosCVPR16}), the vast majority of applications only have limited data in any domain. Finally, most of the above works assume that the target label set matches with the source one, which is often not the case in practice. For example, road scene segmentation across different countries, or segmentation across outdoor and indoor scenes, have domain-specific label sets.

In this paper, we propose and address the novel problem of {\em universal semi-supervised semantic segmentation} as a practical setting for many real-world applications. It seeks to aggregate knowledge from several different domains during training, each of which has few labeled examples but several unlabeled examples. The goal is to simultaneously limit training cost through reduced annotations and deployment cost by obtaining a single model to be used across domains. Label spaces may be partially or fully non-overlapping across domains. While fine-tuning a source model on a small amount of target data is a possible counterpoint, it usually requires plentiful source labels and necessitates deployment of a separate model in every domain due to catastrophic forgetting~\cite{mccloskey1989catastrophic}. Another option is joint training, which does yield a unified model across domains, but does not leverage unlabeled data available in each domain. Our semi-supervised universal segmentation approach leverages both limited labeled and larger-scale unlabeled data in every domain, to obtain a single model that performs well across domains. \tabref{tab:context-table} presents the advantage of the proposed semi-supervised universal segmentation over some of the existing approaches.

In particular, we use the labeled examples in each domain to supervise the universal model, akin to multi-tasking \cite{kokkinos2017ubernet,mccann2018natural,kaiser2017one}, albeit with limited labels. We simultaneously make use of the large number of unlabeled examples to align pixel level deep feature representations from multiple domains using entropy regularization based objective functions. Entropy regularization uses unsupervised examples and helps in encouraging low density separation between the feature representations and improve the confidence of predictions.
Moreover models trained on one domain typically result in noisy predictions and high entropy output maps when deployed in a different domain, and the proposed cross dataset entropy minimization encourages refined prediction maps across datasets.
We calculate the similarity score vector between the encoder outputs at each pixel and the label embeddings (computed from class prototypes~\cite{snell2017prototypical}), and minimize the entropy of this discrete distribution to positively align similar examples between the labeled and the unlabeled images. We do this unsupervised alignment both within domain, as well as across domains.

We believe such within and cross-domain alignment is fruitful even with non-overlapping label spaces, particularly so for semantic segmentation, since label definitions often encode relationships that may positively reinforce performance in each domain. For instance, two road scene datasets such as Cityscapes \cite{cordts2016cityscapes} and IDD \cite{anue} might have different labels, but share similar label hierarchies. Even an outdoor dataset like Cityscapes and an indoor one like SUN \cite{song2015sun} may have label relationships, for example, between horizontal (road, floor) and vertical (building, wall) classes. Similar observations have been made for multi-task training \cite{zamir2018taskonomy}.

We posit that our pixel wise entropy-based objective discovers such alignments to improve over joint training, as demonstrated quantitatively and qualitatively in our experiments. Specifically, our experiments lend insights across various notions of domain gaps. With Cityscapes \cite{cordts2016cityscapes} as one of domains (road scenes in Germany), we derive universal models with respect to CamVid (roads in England) \cite{brostow2009semantic}, IDD (roads in India) \cite{anue} and SUN (indoor rooms) \cite{song2015sun}. In each case, our semi-supervised universal model improves over fine-tuning and joint training, with visualizations of the learned feature representations demonstrating conceptually meaningful alignments. We use dilated residual networks in our experiments \cite{yu2017dilated}, but the framework is equally applicable to any of the existing deep encoder-decoder architectures for semantic segmentation.

\begin{table}[!t]
  \centering
  \Large
  \resizebox{0.5\textwidth}{!}{
  \begin{tabular}{@{} >{\raggedright}p{5.3cm} >{\raggedright}p{2.2cm} >{\raggedright}p{2.2cm} >{\raggedright}p{2cm} p{2cm}}
    \toprule  \\[-1em]
    & Source Unlabeled Data & Target Unlabeled Data 
    & Joint Model & Mixed Labels Support \\
    \midrule
    Fine Tuning & \cross & \cross & \cross & \tick \\
    Semi-supervised \cite{hung2018adversarial, Souly_2017_ICCV} & \tick & \cross & \cross & NA \\
    CyCADA \cite{pmlr-v80-hoffman18a} & \cross & \tick & \tick & \cross \\
    Joint Training & \cross & \cross & \tick & \tick \\
    Our Approach & \tick & \tick & \tick & \tick \\
    
    \bottomrule \\[-1.5ex]
  \end{tabular}
  }
  \vspace{-1em}
  \captionsetup{width=0.45\textwidth}
  \caption{\label{tab:context-table} Comparison of  Universal Semi-Supervised Segmentation against existing methods.}
  \vspace{-1em}
\end{table}

\paragraph{Our Contributions}
\vspace{-0.2cm}
\begin{tight_itemize}
    \item We propose a universal segmentation framework to train a single joint model on multiple domains with disparate label spaces to improve performance on each domain. This framework adds no extra parameters or significant overhead during inference compared to existing methods for deep semantic segmentation.
    
    \item We introduce a pixel-level entropy regularization scheme to train semantic segmentation architectures using datasets with few labeled examples and larger quantities of unlabeled examples (\secref{sec:problem_description}). 
    
    \item We demonstrate the effectiveness of our alignment over a wide variety of indoor~\cite{song2015sun} and outdoor~\cite{cordts2016cityscapes,anue,brostow2009semantic} segmentation datasets with various degrees of label overlaps. We also compare our results with other semi-supervised approaches, based on adversarial losses, giving improved results (\secref{sec:results}). 
    
\end{tight_itemize}

\section{Related Work}
\label{sec:related}


\paragraph{Semantic Segmentation} Semantic segmentation in computer vision is the task of assigning semantic labels to each pixel of an image. Most of the state of the art models for semantic segmentation~\cite{yu2017dilated, long2015fully, noh2015learning, badrinarayanan2015segnet, chen2018deeplab, ERFNet-1} have been possible largely due to breakthroughs in deep learning that have pushed the boundaries of performance in image classification~\cite{krizhevsky2012imagenet, he2015delving, resnets} and related tasks. The pioneering work in~\cite{long2015fully} proposes an end-to-end trainable network for semantic segmentation by replacing the fully connected layers of pretrained AlexNet~\cite{krizhevsky2012imagenet} and VGG Net~\cite{simonyan2014very} with fully convolutional layers that aggregate spatial information across various resolutions. Noh \etal~\cite{noh2015learning} use transpose convolutions to build a learnable decoder module, while DeepLab network~\cite{chen2018deeplab} uses artrous convolutions along with artrous spatial pyramid pooling for better aggregation of spatial features. Segmentation architectures based on dilated convolutions~\cite{yu2015multi} for real time semantic segmentation have also been proposed~\cite{yu2017dilated, ERFNet-1}. 

\begin{figure*}[!t]
\centering
   \includegraphics[width=0.9\textwidth]{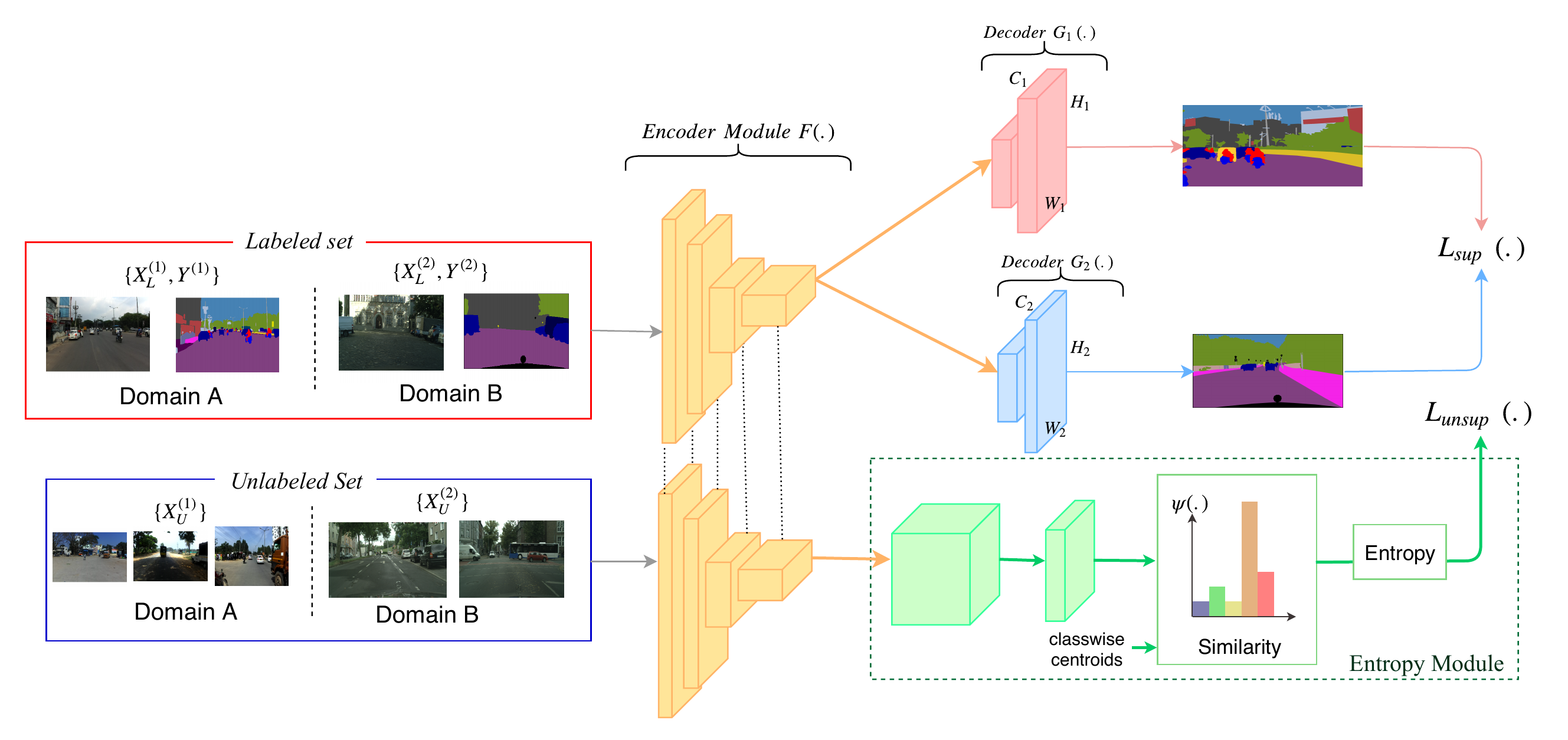}
\vspace{-1em}
\captionsetup{width=\textwidth}
   \caption{Different modules in the proposed universal semantic segmentation framework. $\{\X_l^{(1)} , \Y^{(1)}\}$ , $\{\X_l^{(2)} , \Y^{(2)}\}$ are the set of labeled examples and $\X_u^{(1)}$ , $\X_u^{(2)}$ are the set of unlabeled examples.
   The entropy module uses the unlabeled examples to perform alignment of pixel wise features from multiple domains by calculating pixel wise similarity with the labels, and minimizing the entropy of this discrete distribution.}
\label{fig:model_pic}
\vspace{-1.3em}
\end{figure*}

\paragraph{Semi Supervised Learning} Most of the existing semantic segmentation architectures require large scale annotation of labeled data for achieving good results. To address this limitation, various semi supervised learning methods have been proposed in \cite{Souly_2017_ICCV, papandreou2015weakly, hung2018adversarial, hong2015decoupled, wei2018revisiting}, which make use of easily available large scale unsupervised or weakly supervised data during training. While these approaches deliver competitive results when trained and deployed on a specific dataset, the need for learning efficient segmentation models transferable across domains and environments having limited training data remains. 


\paragraph{Transfer Learning and Domain Adaptation} Transfer learning~\cite{weiss2016survey} involves transferring deep feature representations learned in one domain or task to another domain or task where labeled data availability is low. Previous works demonstrate transfer learning capabilities between related tasks~\cite{donahue2013decaf, zeiler2014visualizing, oquab2014learning, redmon2016you} or even completely different tasks~\cite{girshick2014rich, ren2015faster, long2015fully}. Unsupervised domain adaptation is a related paradigm which leverages labeled data from a source domain to learn a classifier for a new unsupervised target domain in the presence of a domain shift. Various generative and discriminative domain adaptation methods have been proposed for classification tasks in \cite{ganin2014unsupervised, ganin2016domain, Tzeng_2017_CVPR, tzeng2015simultaneous, pinheiro2018unsupervised, Cao_2018_ECCV} and for semantic scene segmentation in \cite{hoffman2016fcns, tsai2018learning, chen2017no, pmlr-v80-hoffman18a, chen2018road, hong2018conditional, zhang2019curriculum}. 

Most of these works in domain adaptation assume equal source and target dataset label spaces or a subset target label space, which is not the most general case for real world applications. To address this limitation with the domain adaptation approaches, we propose a method similar to \cite{luo2017label} which works in the extreme case of non-intersecting label spaces. Moreover, pixel level adaptation based methods are typically focused on using knowledge from a large labeled source domain (eg. Synthia \cite{RosCVPR16}) to improve performance on a specific target domain, while we propose a joint training framework to train a single model that delivers good performance on both the domains.

\paragraph{Universal Segmentation} Multitask learning~\cite{caruana1997multitask} is shown to improve performance for many tasks that share useful relationships between them in computer vision~\cite{sermanet2013overfeat, kokkinos2017ubernet, zamir2018taskonomy}, natural language processing~\cite{collobert2008unified, mccann2018natural, kaiser2017one} and speech recognition~\cite{seltzer2013multi}. Universal Segmentation builds on this idea by training a single joint model that is useful across multiple semantic segmentation domains with possibly different label spaces to make use of transferable representations at lower levels of the network. Liang~\etal~\cite{liang2018dynamic} first propose the idea of universal segmentation by performing dynamic propagation through a label hierarchy graph constructed from an external knowledge source like WordNet. We propose an alternative method to perform universal segmentation without the need for any outside knowledge source or additional model parameters during inference, and instead make efficient use of the large set of unlabeled examples in each of the domains for unsupervised feature alignment. Following the success of metric learning based approaches in tasks such as fine grained classification~\cite{akata2015evaluation, akata2013}, latent hierarchy learning \cite{saha2018} and zero-shot prediction~\cite{norouzi2013zero, frome2013devise, lampert_zeroshot}, we use pixel level class prototypes~\cite{snell2017prototypical} for performing semantic transfer across domains.


\section{Problem Description}
\label{sec:problem_description}

In this section, we explain the framework used to train a single model across different segmentation datasets with possibly disparate label spaces using a novel pixel aware entropy regularization objective. 

We have $d$ datasets $\{\D^{(i)}\}_{i=1}^d$, each of which has few labeled examples and many unlabeled examples.
The labeled images and corresponding labels from $\D^{(i)}$ are denoted by $ \{\X_l^{(i)} , \Y^{(i)}\}_{i=1}^{N_l^{(i)}}$, where $\Y^{(i)} \in \LY_i$, and $N_l^{(i)}$ is the number of labeled examples. The unlabeled images are represented by $\{\X_u^{(i)}\}_{i=1}^{N_u^{(i)}}$, and $N_u^{(i)}$ is the number of unlabeled examples. We work with domains with very few labeled examples ($N_u^{(i)} \gg  N_l^{(i)}$), and consider the general case of non-intersecting label spaces, so that $\LY_p \neq \LY_q$ for any $ p,q $. The label spaces might still have a partial overlap between them, which is common in the case of segmentation datasets. For ease of notation, we consider the special case of two datasets $\{\D^{(1)}, \D^{(2)}\}$, but similar idea can be applied for the case of multiple datasets as well.

The proposed universal segmentation model is summarized in \figref{fig:model_pic}. 
Deep semantic segmentation architectures generally consist of an encoder module which aggregates the spatial information across various resolutions and a decoder module that consists of a classifier and an up sampler to enable pixel wise predictions at a resolution that matches the input. In order to enable joint training with multiple datasets, we modify this encoder decoder architecture by having a shared encoder module $\mathcal{F}$ and different decoder layers $\mathcal{G}_1(.)$, $\mathcal{G}_2(.)$ for prediction in different label spaces. For a labeled input image $\x_l$, the pixel wise predictions are denoted by $\hat{y}^{(k)} = \mathcal{G}_k(\mathcal{F}(\x_l))$ for $k={1,2}$ which, along with the labeled annotations, gives us the supervised loss. To make use of the semantic information from the unlabeled examples $\X_u $, we propose an entropy regularization module $\mathcal{E}$. This entropy module takes as input the output of the encoder $\mathcal{F}(.)$ to give pixel wise representation outputs in an embedding space. The entropy of the similarity score vector of these embedding representations with the label embeddings results in the unsupervised loss term. Each of these loss terms is explained in detail in the following sections. 

\paragraph{Supervised Loss}
\label{subsec:supervised}
The supervised loss is 
the softmax cross entropy loss between the predicted segmentation mask $ \hat{y}$ and the corresponding pixel wise ground truth segmentation masks for all labeled examples. Specifically, for the samples from dataset $k$, 
\begin{equation}
    \LS_{S}^{(k)} = \frac{1}{N_{l}^{(k)}} \sum_{\x_i \in \D^{(k)}} \psi_k\left(\y_i , \mathcal{G}_k\left(\mathcal{F}\left(\x_i\right)\right)\right),
    \label{eq:sup_loss}
\end{equation}

where $\psi_k$ is the softmax cross entropy loss function over the label space $\LY_k$, which is averaged over all the pixels of the segmentation map. $\LS_{S}^{(1)}$ and  $\LS_{S}^{(2)}$ together comprise the supervised loss term $\LS_{S}$. 

\paragraph{Entropy Module}
\label{subsec:ent_mod}

The large number of unsupervised images available provides us with rich information regarding the visual similarity between the domains and the label structure, which the existing methods on adversarial based semi supervised segmentation~\cite{Souly_2017_ICCV, hung2018adversarial} or universal segmentation~\cite{liang2018dynamic} do not exploit. To address this limitation, we propose using entropy regularization to transfer the information from labeled images to the unlabeled images, as well as among the unlabeled images between the datasets. Entropy regularization is proved to encourage low density separation between the clusters in the feature space~\cite{grandvalet2005semi}, hence resulting in high confidence predictions and smooth output maps for semi supervised learning. A crucial difference between some previous works which use entropy regularization for semi supervised learning~\cite{grandvalet2005semi, long2016unsupervised, vu2018advent} and ours is that we perform entropy regularization in a separate embedding space using an \textit{entropy module} $\mathcal{E}$, unlike the other works which apply this objective directly in the softmax output space. This embedding approach helps in achieving semantic transfer between datasets with disparate label sets, hence aiding in closely aligning the visually similar pixel level features calculated from the segmentation network from both the datasets.

\begin{figure}[!t]
\centering
   \includegraphics[width=0.6\linewidth]{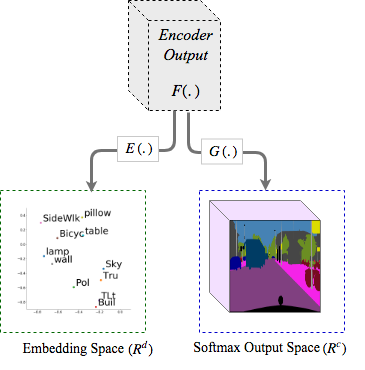}
\vspace{-1em}
\captionsetup{width=0.45\textwidth}
   \caption{
   In addition to a traditional decoder layer that outputs predictions in the respective label spaces $\mathbb{R}^c$, we also have an entropy module $\mathcal{E}(.)$ that first maps the features of both the domains into a common embedding space $\mathbb{R}^d$, and then calculates similarity scores with the label embeddings of respective datasets.}
\label{fig:maping_pic}
\vspace{-1.3em}
\end{figure}

The entropy module is explained in~\figref{fig:maping_pic}, and works similar to the decoder module in a segmentation architecture. Firstly, we project the encoder outputs from the segmentation network from both datasets into a common $d$ dimensional embedding space $\mathbb{R}^d$, and upsample this output map to match the size of the input.
Then, a similarity metric $\phi$, which operates on each pixel, is used to calculate the similarity score of the embedding representations with each of the $d$ dimensional label embeddings using the equation
\begin{equation}
    \left[v_{ij}\right]_k = \phi\left( \mathcal{E} \left(\mathcal{F}\left(\x_u^{(i)}\right)\right) , \mathrm{c}_k^{(j)}\right) \;\; \forall k \in \{|\LY_j|\},
    \label{eq:sim_scores}
\end{equation}
where $\x_u^{(i)}$ is an image from the $i^{\text{th}}$ unlabeled set, $\mathrm{c}_k^{(j)} \in \mathbb{R}^d$ is the label embedding corresponding to the $k^{\text{th}}$ label from the $j^{\text{th}}$ dataset and $[v_{ij}] \in \mathbb{R}^{|\LY_j|}$. When $i = j$, the scores correspond to the similarity scores within a dataset, and when $i \neq j$, they provide the cross dataset similarity scores. The label embeddings are just the prototype features calculated using the labeled data. They are pre computed and kept fixed over the course of training the network, since we found that the limited supervised data was not sufficient to jointly train a universal segmentation model as well as fine tune the label embeddings. More details on calculating label embeddings is presented in the supplementary section. 


\paragraph{Unsupervised Loss}
\label{subsec:unsup_loss}

We have two parts for the unsupervised entropy loss. 
The first part, the cross dataset entropy loss, is obtained by minimizing the entropy of the cross dataset similarity vectors.
\begin{equation}
    \LS_{US,c} = \mathcal{H}(\sigma([v_{12}])) + \mathcal{H}(\sigma([v_{21}])),
    \label{eq:cross_entropy}
\end{equation}
where $\mathcal{H}(.)$ is the entropy measure of a discrete distribution, $\sigma(.)$ is the softmax operator and the similarity vector $[v]$ is from~\eqref{eq:sim_scores}. Minimizing $\LS_{US,c}$ makes the probability distribution \textit{peaky} over a single label from a dataset, which helps in label side semantic transfer across datasets and hence improving the overall prediction certainty of the network.
In addition, we also have a within dataset entropy loss given by
\begin{equation}
    \LS_{US,w} = \mathcal{H}(\sigma([v_{11}])) + \mathcal{H}(\sigma([v_{22}]))
    \label{eq:within_entropy}
\end{equation}
which aligns the unlabeled examples within the same domain. 

The total loss $\LS_T$ is the sum of the supervised loss from~\eqref{eq:sup_loss}, and the unsupervised losses from \eqref{eq:cross_entropy} and \eqref{eq:within_entropy}, written as 
\begin{eqnarray}
    \LS_T =& \LS_{S}(\X_l^{(1)} , \Y^{(1)} , \X_l^{(2)} , \Y^{(2)}) 
             + \alpha \cdot \LS_{US,c}(\X_u^{(1)} , \X_u^{(2)}) \nonumber \\
             &+ \beta \cdot \LS_{US,w}(\X_u^{(1)} , \X_u^{(2)})
    \label{eq:total_loss}
\end{eqnarray}
where $\alpha$ and $\beta$ are a hyper parameters that control the influence of the unsupervised loss in the total loss.

\normalsize
\paragraph{Inference} For a query image $\mathrm{q}^{(k)}$ from dataset $k$ during test time, the output $\hat{\y}^{(k)} = \mathcal{G}_k(\mathcal{F}(\mathrm{q}^{(k)}))$ gives us the segmentation map over the label set $\LY_k$ and the pixel wise label predictions. This adds no computation overhead or extra parameters to our approach during inference compared to existing deep semantic segmentation approaches. We note that although we calculate feature and label embeddings in our method and metric based inference schemes like nearest neighbor search might enable prediction in a label set agnostic manner, calculating pixel wise nearest neighbors using existing methods can prove very slow and costly for images with high resolution. 


\begin{table*}[!t]
\centering

\resizebox{0.96\textwidth}{!}{
\begin{tabular}{@{} l *{19}{c} c c @{}}
    \toprule  \\[-1em]
    Method & 
    \rotatebox{0}{\footnotesize Road} & \rotatebox{0}{\footnotesize SideWalk} & \rotatebox{0}{\footnotesize Building} & \rotatebox{0}{\footnotesize Wall} & \rotatebox{0}{\footnotesize Fence} & \rotatebox{0}{\footnotesize Pole} & \rotatebox{0}{\footnotesize Traff. lt.} & \rotatebox{0}{\footnotesize Traff. Sgn.} & \rotatebox{0}{\footnotesize Veg.} & \rotatebox{0}{\footnotesize Train} & \rotatebox{0}{\footnotesize Sky} & \rotatebox{0}{\footnotesize Person} & \rotatebox{0}{\footnotesize Rider} & \rotatebox{0}{\footnotesize Car} & \rotatebox{0}{\footnotesize Truck} & \rotatebox{0}{\footnotesize Bus} & \rotatebox{0}{\footnotesize Train} & \rotatebox{0}{\footnotesize MotorCyc.} & \rotatebox{0}{\footnotesize Bicycle} && 
    \rotatebox{0}{ mIoU} \\ 
    \midrule \\[-1.5ex]
    CS only & 
    91.76 & \textbf{54.78} & 80.02 & 3.70 & 16.58 & 29.84 & 22.31 & 33.74 & \textbf{83.88} & 32.89 & \textbf{82.07} & 52.67 & \textbf{21.57} & 81.11 & \textbf{19.01} & 3.87 & 0.0 & \textbf{19.64} & \textbf{49.01 }
    && 40.97 \\
    
    Univ-basic & 
    87.00 & 44.54 & 77.77 & \textbf{10.21} & 11.07 & 25.54 & 14.51 & 25.82 & 80.72 & 22.40 & 78.19 & 49.00 & 19.64 & 75.35 & 1.86 & 0.25 & \textbf{10.98} & 8.83 & 41.08 
     && 36.04 \\
   
    Univ-full & 
    \textbf{92.18} & 51.29 & \textbf{80.07} & 0.0 & \textbf{24.01} & \textbf{33.73} & \textbf{26.16} & \textbf{38.71} & 82.30 & \textbf{36.39} & 81.61 & \textbf{54.38} & 20.48 & \textbf{81.71} & 2.37 & \textbf{22.79} & 3.85 & 1.31 & 46.23 
     && \textbf{41.03} \\
    
   \bottomrule
\end{tabular}}

\vspace{0.1em}
\small
\resizebox{0.96\textwidth}{!}{
\begin{tabular}{@{} l*{11}{c} c c @{}}
    \toprule  \\[-1em]
    Method & 
    \rotatebox{0}{\footnotesize Sky} &
    \rotatebox{0}{\footnotesize Buil.} & \rotatebox{0}{\footnotesize Pole} & \rotatebox{0}{\footnotesize Road} & \rotatebox{0}{\footnotesize Pave.} & \rotatebox{0}{\footnotesize Tree} & \rotatebox{0}{\footnotesize Sign} & \rotatebox{0}{\footnotesize Fence} & \rotatebox{0}{\footnotesize Car} & \rotatebox{0}{\footnotesize Ped.} & \rotatebox{0}{\footnotesize Bicy.} && \rotatebox{0}{\footnotesize mIoU} \\ 
    \midrule \\[-1.5ex]
    
    Camvid only & 
     85.58 &  75.15 &   8.17 &  84.86 &  52.34 &  69.68 &  27.11 &  20.48 &  73.1 &   24.36 &  29.42
     && 50.02 \\
    
    Univ-basic &
    \textbf{87.04} &  76.67 &   9.56 &  83.5 &   51.35 &  70.07 &  27.75 &  22.6 &   \textbf{73.22} &  33.94 &  35.25
     && 51.9 \\
    
    Univ-full & 
    86.3 &   \textbf{77.23} &  \textbf{17.13} &  \textbf{84.99} & \textbf{ 53.35} &  \textbf{70.57} &  \textbf{31.99} &  \textbf{32.45} &  72.94 &  \textbf{36.61} &  \textbf{37.22}
     && \textbf{54.62} \\
    
    \bottomrule \\[-1.5ex]
    
\end{tabular}}
\vspace{-1em}
\captionsetup{width=0.95\textwidth}
\caption{Class-wise IoU values for the 19 classes in Cityscapes dataset and 11 classes in the CamVid dataset with various ablations of universal semantic segmentation models, for N=100 on Resnet-18. Note the improvement of our method (Univ-full) for smaller classes like \textit{pole} and \textit{sign} on Cityscapes and CamVid datasets.}
\label{tab:class_wise_iou}
\vspace{-1em}
\end{table*}



 \section{Experiments and Results}
\label{sec:results}

\normalsize


We provide the performance results of the proposed approach on a wide variety of real world datasets used in autonomous driving as well as indoor segmentation settings. We show the superiority of the our method over the existing baselines (\secref{subsec:ablation}), demonstrate improvement upon the state of the art semi-supervised approaches (\secref{subsec:SOA}), and also show the results on cross domain datasets (\secref{secref:cross_domain}). Using only a fraction of the labeled data available, we show competitive results on these datasets.



\begin{table}
  \centering
  \resizebox{0.45\textwidth}{!}{
  \begin{tabular}{@{} l c c c c c c c @{}} 
    \toprule  \\[-1em]
     \multirow{2}{*}{Method} &
     \multicolumn{3}{c}{N=50}  && \multicolumn{3}{c}{N=100} \\ 
      \cmidrule{2-4} \cmidrule{6-8} 
    & CS &  CVD & Avg. && CS & CVD & Avg. \\
    \midrule \\[-1.5ex]
    Train on CS & 33.33 & 32.92 & 33.13 && 40.97 & 36.52 & 38.75 \\ 
    Train on CVD & 19.47 & 42.81 & 31.14 && 22.20 & 50.02 & 36.11 \\ 
    \midrule
    Univ-basic ($\LS_s$) & 32.82 & 48.56 & 40.69 && 36.04 & 51.90 & 43.97 \\ 
    Univ-cross (+ $\LS_c$) & 33.86 & 52.57 & 43.22 && 37.82 & 49.31 & 43.57 \\
    Univ-full (+ $\LS_c,\LS_w$) & 34.01 & 53.23 & \textbf{43.62} && 41.03 & 54.62 & \textbf{47.83} \\ 
    \bottomrule \\[-1.5ex]
  \end{tabular}}
 \vspace{-0.5em}
  \caption{\label{tab:few_shot_cs_cvd} mIoU values for universal segmentation using Cityscapes (CS) and CamVid (CVD) datasets with a Resnet-18 backbone. $N$ is the number of supervised examples available from each dataset. Bold entries have the highest \textit{average mIoU} across the datasets.}
  \vspace{-1.2em}
\end{table}

\subsection{Training Details}
\label{subsec:exp_details}

\paragraph{Datasets} We show the results of our approach on large scale urban driving datasets from various domains like Cityscapes \cite{cordts2016cityscapes} (CS), CamVid \cite{brostow2009semantic} (CVD) and Indian Driving Dataset (IDD) \cite{autonue, anue}. 

Cityscapes~\cite{cordts2016cityscapes} is a standard autonomous driving dataset consisting of 2975 training images collected from various cities across Europe finely annotated with 19 classes. CamVid \cite{brostow2009semantic} dataset contains 367 training, 101 validation and 233 testing images taken from video sequences finely labeled with 32 classes, although we use the more popular 11 class version from~\cite{badrinarayanan2015segnet}. We also demonstrate results on IDD~\cite{autonue,anue} dataset, which is an in-the-wild dataset for autonomous navigation in unconstrained environments. It consists of 6993 training and 981 validation images finely annotated with 26 classes collected from 182 drive sequences on Indian roads, taken in highly varying weather and environment conditions. This is a challenging driving dataset since it contains images taken from largely unstructured environments. 

While these autonomous driving datasets typically offer many challenges, there is still limited variation with respect to the classes, object orientation or camera angles. Therefore, we also use SUN RGB-D~\cite{song2015sun} dataset for indoor segmentation, which contains 5285 training images along with 5050 validation images finely annotated with 37 labels consisting of regular household objects like chair, table, desk, pillow etc. We report results on the 13 class version used in~\cite{handa2016understanding}, and use only the RGB information for our universal training and ignore the depth information provided. 

\vspace{-1em}
\paragraph{Architecture}
\normalsize
Although the proposed framework is readily applicable to any state-of-the art encoder-decoder semantic segmentation framework, we use the openly available PyTorch implementation of dilated residual network~\cite{yu2017dilated} owing to its low latency in autonomous driving applications.
%
We take the embedding dimension $d$ to be $128$, and use dot product for the pixel level similarity metric $\phi(.)$ as it can be implemented as a $1 \times 1$ convolution on most of the modern deep learning packages. More details for each experimental setting is presented in the supplementary section.

\vspace{-1em}
\paragraph{Evaluation Metric} We use the mean IoU (Intersection over Union) as the performance analysis metric. The IoU for each class is given by
\begin{equation}
    \text{IoU} = \frac{\text{TP}}{\text{TP+FP+FN}},
\end{equation}
where TP , FP , FN are the true positive, false positive and false negative pixels respectively, and mIoU is the mean of IoUs over all the classes. mIoUs are calculated separately for all datasets in a universal model. All the mIoU values reported are on the publicly available validation sets for the CS, IDD and SUN-RGB datasets, and on the test set for the CamVid dataset.


\subsection{Ablation Studies}
\label{subsec:ablation}

We perform the following ablation studies in our experiments to provide insights into the various components of the proposed objective function. (i) \textit{Train on source}: We train a semantic segmentation network using \textit{only} the limited training data available on one dataset, and provide results when tested on both the datasets. Since the label spaces do not directly overlap, we finetune a different classifier (decoder) for both the datasets and keep the feature extractor (encoder) as the same.
%
\begin{table}[!t]
  \centering
  \resizebox{0.3\textwidth}{!}{
  \begin{tabular}{@{} l c c c @{}}
    \toprule  \\[-1em]
    \multirow{2}{*}{Method} & 
    \multicolumn{2}{c}{N=375} \\
    \cmidrule{2-4}
    & CS & CamVid & Avg. \\
    \midrule \\[-1.5ex]
    Train on CS & 55.07 & 48.52 & 51.80 \\ 
    Train on CVD & 26.45 & 60.61 & 43.53 \\
    Hung \etal~\cite{hung2018adversarial} & 58.80 & - & - \\
    Souly \etal~\cite{Souly_2017_ICCV} & - & 58.20 & - \\
    \midrule
    Univ-basic ($\LS_s$) & 53.14 & 65.33 & 59.24 \\ 
    Univ-cross (+ $\LS_c$) & 56.36 & 63.34 & 59.85 \\
    Univ-full (+ $\LS_c,\LS_w$) & 55.92 & 64.72 & \textbf{60.32} \\ 
    \bottomrule \\[-1.5ex]
  \end{tabular}}
  \vspace{-0.5em}
  \captionsetup{width=0.46\textwidth}
  \caption{\label{tab:CVD_resnet-101} Comparison of our approach with other semi-supervised approaches on the Resnet-101 backbone and CS+CVD dataset. Our approach (Univ-full) results in a single model across datasets unlike the previous semi-supervised approaches and deliver competitive performance on both the datasets.
}
  \vspace{-1.5em}
\end{table}
%
(ii) \textit{Univ-basic}: To study the effect of the unsupervised losses, we put $\alpha,\beta=0$ and perform training using only the supervised loss term from~\eqref{eq:sup_loss} and no entropy module at all. This is similar to plain joint training using the supervised data from each domain. (iii) \textit{Univ-cross}: To study the effect of the cross dataset loss term from \eqref{eq:cross_entropy}, we conduct experiments by adding $\alpha=1$ to the loss term. (iv) \textit{Univ-full}: This is the proposed model, including all the supervised and unsupervised loss terms. We use $\alpha , \beta = 1$ in the loss function in~\eqref{eq:total_loss}. The best model is defined as the model having the highest \textit{average mIoU} across the datasets.

Although many works on domain adaptation also provide results on Cityscapes dataset, we note that we cannot directly compare our result against them, since the problem setting is very different. While most of the domain adaptation approaches use large scale synthetic datasets as source dataset to improve performance on a specific target domain, we train our models on multiple resource constrained real world datasets directly. 

\begin{table}[!t]
  \centering
  \resizebox{0.45\textwidth}{!}{
  \begin{tabular}{@{} l c c c c c c c @{}}
    \toprule  \\[-1em]
    \multirow{2}{*}{Method} & 
    \multicolumn{3}{c}{N=100} && \multicolumn{3}{c}{N=1500} \\
    & \multicolumn{3}{c}{(Resnet-18)} && \multicolumn{3}{c}{(Resnet-50)} \\
    \cmidrule{2-4} \cmidrule{6-8}
    & CS & IDD & Avg. && CS & IDD & Avg.  \\
    \midrule \\[-1.5ex]
    Train on CS & 40.97  & 14.64 & 27.81 && 64.23 & 32.50 & 48.37 \\ 
    Train on IDD & 25.05 & 26.53 & 25.79 && 46.32 & 55.01 & 50.67 \\ 
    \midrule
    Univ-basic  & 37.94 & 25.21 & 31.58 && 63.55 & 53.21 & 58.38 \\ 
    Univ-full & 36.48 & 27.45 & \textbf{31.97} && 64.12 & 55.14 & \textbf{59.63} \\
    \bottomrule \\[-1.5ex]
  \end{tabular}}
  \vspace{-0.5em}
  \caption{\label{tab:few_shot_anue_cs} Universal segmentation results using IDD and CS datasets. Our approach (Univ-full) performs better across Resnet-18 and Resnet-50 CNN backbones.}
  \vspace{-1.2em}
\end{table}

\vspace{-0.5em}

\paragraph{Cityscapes + CamVid} The results for training a universal model on Cityscapes and CamVid datasets is given in \tabref{tab:few_shot_cs_cvd}. For a setting of N=100 which corresponds to using 100 labeled examples from each domain, the proposed method gives the best mIoU value of $41.03\%$ on Cityscapes and $54.62\%$ on CamVid clearly outperforming the baseline approaches. Moreover, the universal segmentation method using the proposed unsupervised losses also performs better than using only supervised losses, which demonstrates the advantage of having unsupervised entropy regularization in domains with few labeled data and lots of unlabeled data.

Another observation from \tabref{tab:few_shot_cs_cvd} is that for N=100, a model trained only on Cityscapes suffers a performance drop of ~13.5\% mIoU when tested on the CamVid dataset compared to a model trained on Camvid alone. Similarly, the performance drop in the case of Cityscapes is ~18\% mIoU for a model trained on Camvid. Therefore, it is evident that models trained on one dataset, like Cityscapes do not always perform well when deployed on a different dataset, like CamVid, due to domain shift and result in noisy predictions and poor output maps. This further brings out the necessity of training a single model which performs well on both the domains by using an entropy regularization based semantic transfer objective.

In the case of semantic segmentation datasets, very low values of N offers challenges like limited representation for many of the smaller labels, but we notice that the proposed model for N=50 still manages to perform consistently better on both the datasets.

Comparison of class-wise mIoUs of the universal segmentation approach for N=100 with CS+CamVid is given in~\tabref{tab:class_wise_iou}. Entropy regularization clearly boosts performance in 9 of the 11 classes on the CamVid dataset, and for 10 out of 19 classes on the Cityscapes dataset. More importantly, it is the smaller classes like \textit{pole}, \textit{traffic sign}, \textit{pedestrian} and \textit{fence} which benefit greatly from universal training on both the Cityscapes and CamVid datasets, in spite of using only a small fraction of the labeled examples from these datasets.

\vspace{-0.5em}
\paragraph{IDD + Cityscapes}  This combination is a chosen for validating the universal segmentation approach as the images are from widely dissimilar domains in terms of geography, weather conditions as well as traffic setup, and the datasets together capture the wide variety of road scenes one might encounter while training autonomous driving datasets for vision based navigation. The results for universal semantic segmentation using IDD and Cityscapes (CS) are shown in~\tabref{tab:few_shot_anue_cs}. Using 100 training examples from each domain, the proposed univ-full model gives an mIoU of 36.48\% on Cityscapes (CS) and 27.45\% on IDD using a Resnet-18 backbone, performing better than the univ-basic method on the average mIoU. 

Similar to the CS+CamVid case, the features trained on Cityscapes dataset do not transfer directly to IDD, and shows a performance drop of ~12\% mIoU, demonstrating the necessity of learning universal representations for large scale datasets as well.

Furthermore, as an extreme case, we show the utility of the proposed approach even in the case of large number of labeled examples. We choose N=1500, which is a challenging setting since the number of supervised examples are already sufficient to train a joint model. However, from ~\tabref{tab:few_shot_anue_cs}, the Resnet-50 based universal model still provides advantage over joint training method, which proves that adding unsupervised examples always helps the training, and more unsupervised examples can be added to these datasets to push the state of the art performance.



\begin{figure*}
\centering
        \begin{subfigure}[b]{0.15\textwidth}
                \centering
                \includegraphics[width=\linewidth]{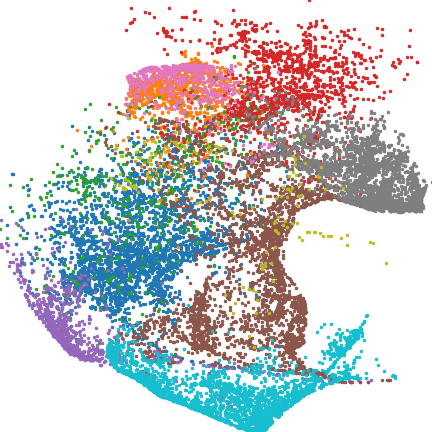}
                \caption{\label{fig:cvd_noentropy}}
                
        \end{subfigure}\hfill
        \begin{subfigure}[b]{0.15\textwidth}
                \centering
                \includegraphics[width=\linewidth]{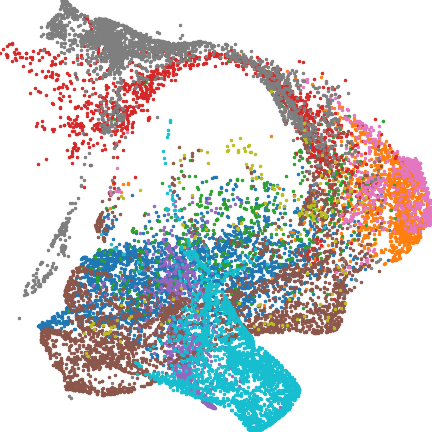}
                \caption{\label{fig:cvd_entropy}}
                
        \end{subfigure}\hfill
        \begin{subfigure}[b]{0.1\textwidth}
                \centering
                \includegraphics[width=\linewidth]{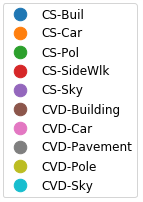}
        \end{subfigure}\hfill
        %
        \begin{subfigure}[b]{0.15\textwidth}
                \centering
                \includegraphics[width=\linewidth]{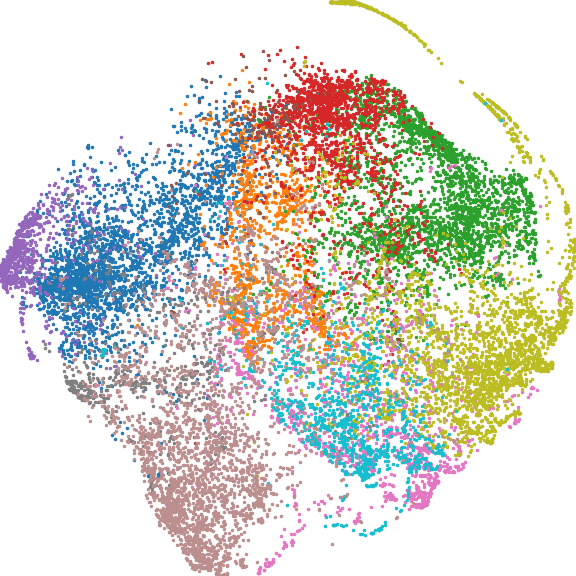}
                \caption{\label{fig:sun_noentropy}}
                
        \end{subfigure}\hfill
        \begin{subfigure}[b]{0.15\textwidth}
                \centering
                \includegraphics[width=\linewidth]{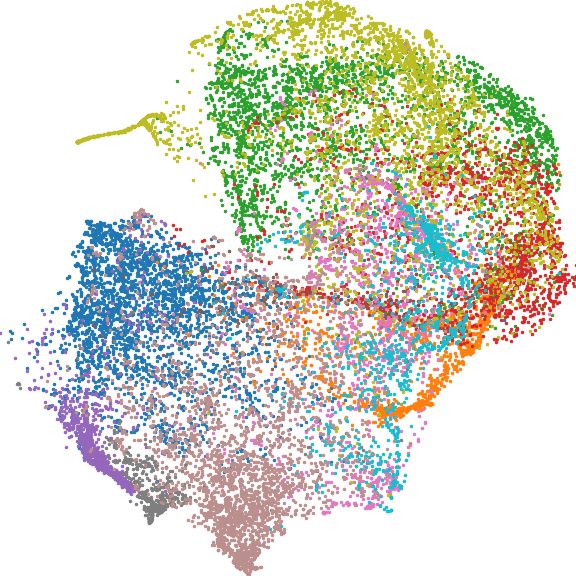}
                \caption{\label{fig:sun_entropy}}
                
        \end{subfigure}\hfill
        \begin{subfigure}[b]{0.1\textwidth}
                \centering
                \includegraphics[width=\linewidth]{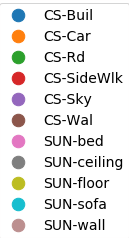}
        \end{subfigure}\hfill
        \smallskip
        \vspace{-1em}
        \captionsetup{width=0.95\textwidth}
        \caption{tSNE visualizations of the encoder output representations for majority classes from CS, CVD and SUN datasets. Plots (\subref{fig:cvd_noentropy}) and (\subref{fig:cvd_entropy}) are for the Univ-basic and Univ-full model from CS-CVD datasets. Observe that the feature embeddings for large classes like \textit{CS:Building-CVD:Building}, \textit{CS:SideWalk-CVD:Pavement}, \textit{CS:Sky-CVD:Sky} align a lot better with universal model. Plots (\subref{fig:sun_noentropy}) and (\subref{fig:sun_entropy}) are for the Univ-basic and Univ-full model from CS-SUN datasets, and labels with similar visual features like \textit{CS:Road - SUN:Floor } show better feature alignment. Best viewed in color and zoom.}
        \label{fig:label_tsne}
        \vspace{-1.2em}
\end{figure*}


\subsection{Comparison with state-of-the art}
\label{subsec:SOA}

In addition to demonstrating the superiority of the proposed method over the baseline approaches, we also compare some of the existing semi supervised semantic segmentation works (which are targeted towards single dataset) with ours in ~\tabref{tab:CVD_resnet-101}, for similar amounts of labeled training data.  
Our model which uses dilated residual network gives competitive results on Cityscapes validation set when compared to \cite{hung2018adversarial} which uses a more complex DeepLab-V2 architecture. Similarly, without using any unsupervised video images unlike \cite{Souly_2017_ICCV}, we show superior results on the CamVid test set compared to them, in spite of the fact that our model is trained to perform well on multiple datasets at once. Most of the previous works optimize adversarial losses, and our results prove that entropy minimization is better suited for semi supervised approaches where limited supervision is  available. 

%


\begin{table}[!t]
  \centering
  \resizebox{0.45\textwidth}{!}{
  \begin{tabular}{l p{2cm} c c c c }
    \toprule  \\[-1em]
    Method & Labeled Examples && CS & SUN & Avg.   \\
    \midrule \\[-1.5ex]
    Train on CS & 1.5k && 64.23 &  15.47 & 39.85 \\ 
    Train on SUN & 1.5k && 15.61 & 42.52 & 29.07 \\  
    \midrule
    SceneNet~\cite{mccormac2017scenenet} & Full(5.3k) && - & 49.8 & - \\
    Univ-basic & 1.5k && 58.01 & 31.55 & 44.78 \\
    Ours[Univ-full]  & 1.5k && 57.91 & 43.12  &  \textbf{50.52} \\
    \bottomrule \\[-1.5ex]
    
  \end{tabular}}
  \vspace{-0.5em}
  \caption{\label{tab:cross_task} mIoU values for universal segmentation across different task datasets with Resnet-50 backbone. While Cityscapes is an autonomous driving dataset, SUN dataset is mainly used for indoor segmentation. This demonstrates the effectiveness of universal segmentation even across diverse environments. 
    }
  \vspace{-1.2em}
\end{table}

\subsection{Cross Domain Experiment}
\label{secref:cross_domain}

A useful advantage of the universal segmentation model is its ability to perform knowledge transfer between datasets used in completely different settings, due to its effectiveness in exploiting useful visual relationships. We demonstrate this effect in the case of joint training between Cityscapes, which is a driving dataset with road scenes used for autonomous navigation and SUN RGB-D, which is an indoor segmentation dataset with household objects used for high-level scene understanding. 

The label sets in Cityscapes and SUN-RGBD dataset are completely different (non overlapping), so the simple joint training techniques generally give poor results.
However, from \tabref{tab:cross_task}, our model outperforms the baselines and provides a good joint model across the domains making use of the unlabeled examples. We also compare our work against SceneNet~\cite{mccormac2017scenenet}, which uses large scale synthetic examples with RGB and depth data for pre-training, as well as all of the 5.3k available labeled examples for training. Using \textit{only 28\%} of the training data from the SUN-RGB dataset, and limited supervision from Cityscapes instead of synthetic examples, we achieve upto 88\% of the mIoU reported in ~\cite{mccormac2017scenenet}. 

\subsection{Feature Visualization}

A more intuitive understanding of the feature alignment performed by our universal model is obtained from the tSNE embeddings~\cite{maaten2008visualizing} of the visual features. The pixel wise output of the encoder module is used to plot the tSNE of selected labels in~\figref{fig:label_tsne}. For the universal training between CS and CVD in Figures~\ref{fig:cvd_noentropy} and \ref{fig:cvd_entropy}, we can observe that classes like \textit{Building-CS} and \textit{Building-CVD}, as well as \textit{Sidewalk-CS} and \textit{Pavement-CVD} align with each other better when trained using a universal segmentation objective. For the universal training between CS and SUN from \figref{fig:sun_noentropy} and \figref{fig:sun_entropy}, labels with similar visual attributes such as \textit{Road} and \textit{Floor} align close to each other in spite of the label sets themselves being completely non overlapping.

\section{Conclusion}
\label{sec:conclusion}

In this work, we demonstrate a simple and effective way to perform universal semi-supervised semantic segmentation. We train a joint model using the few labeled examples and large amounts of unlabeled examples from each domain by an entropy regularization based semantic transfer objective. We show this approach to be useful in better alignment of the visual features corresponding to different domains. We demonstrate superior performance of the proposed approach when compared to supervised training or joint training based methods over a wide variety of segmentation datasets with varying degree of label overlap. We hope that our work would address the growing concern in the deep learning community over the difficulty involved in collection of large number of labeled examples for dense prediction tasks such as semantic segmentation. 
We also believe that other computer vision tasks like object detection and instance aware segmentation can benefit greatly from the ideas discussed in this work.


\paragraph{Acknowledgement} M. Chandraker was supported by NSF CAREER 1751365.

{
\bibliographystyle{ieee_fullname}
\bibliography{scripts/egbib}
}

\newpage
\appendix

\section{Training Details}

We give details of the parameters used for training the universal segmentation models in various settings. We use the openly available PyTorch implementation of dilated residual network~\cite{yu2017dilated}, with encoders designed using ResNet-18 (\textit{drn-d-22}), ResNet-50 (\textit{drn-d-54}) as well as ResNet-101 (\textit{drn-d-105}) architectures. The decoder consists of a $1\x1$ convolution layer followed by a bilinear upsampling layer. We train every model on 2 Nvidia GeForce GTX 1080 GPUs for 200 epochs. During training, we use a crop size of $512\x512$ for Cityscapes and IDD datasets, $360\x480$ for the Camvid dataset and $480\x640$ for the SUN-RGB dataset. Validation mIoUs are reported on the standard resolutions from the dataset. We employ SGD learning algorithm with an initial learning rate of $0.001$ and a momentum of $0.9$, along with a poly learning rate schedule with a power of $0.9$~\cite{chen2018deeplab}. We use a batch size of 10, and take the embedding dimension to be $128$. The default values for $\alpha$ and $\beta$ are taken to be 1. 

\section{Label Embeddings}

\subsection{Calculating the label embeddings}

In this section, we describe the method used to obtain the vector representations for the labels. For each dataset separately, we train an end-to-end segmentation network from scratch using only the limited training data available in that dataset. We use this trained segmentation network to calculate the encoder outputs of the training data at each pixel. 
Typically, the size of the output dimension of the encoder at each pixel (512 for a ResNet encoder) is not equal to the dimension of the label embeddings ($d$=128, in our case). So we first apply a dimensionality reduction technique like PCA to reduce the dimension of the outputs to match the dimension of the label embeddings $d$, and then calculate the class wise centroids to obtain the label embeddings.

\subsection{Updating the label embeddings}

\begin{table}
  \centering
  \resizebox{0.45\textwidth}{!}{
  \begin{tabular}{@{} l c c c c c c c @{}} 
    \toprule  \\[-1em]
     \multirow{2}{*}{Method} &
     \multicolumn{3}{c}{N=50}  && \multicolumn{3}{c}{N=100} \\
      \cmidrule{2-4} \cmidrule{6-8}
    & CS &  CVD & Avg. && CS & CVD & Avg. \\
    \midrule \\[-1.5ex]
    Ours[$\theta=0.5$] & 33.28 & 48.7 & 40.99 && 33.51 & 49.49 & 41.50 \\
    \midrule \\[-1.5ex]
    Ours[Word2Vec] & 33.48 & 53.19 & 43.34 && 36.18 & 52.72 & 44.45 \\
    \midrule \\[-1.5ex]
    Ours[K=1] & 34.01 & 53.23 & 43.62 && 41.03 & 54.62 & 47.83 \\ 
    Ours[K=3] & 35.23 & 52.38 & \textbf{43.81} && 41.82 & 54.96 & \textbf{48.39} \\
    Ours[K=5] & 33.76 & 52.77 & 43.27 && 40.08 & 55.02 & 47.55 \\
    \midrule \\[-1.5ex]
    Direct SER & 21.36 & 35.24 & 28.30 && 23.7 & 30.67 & 27.19 \\
    
    \bottomrule \\[-1.5ex]
  \end{tabular}}
 \vspace{-0.5em}
  \caption{\label{tab:table} Extension of Table 3 from the original paper. $\theta$ is the update factor during training, and the default value is 1. K is the number of embeddings per label. \textit{Ours[Word2Vec]} uses word vectors as label embeddings. The model gives best performance for K=3, $\theta=1$ while using prototype embeddings.}
  \vspace{-1em}
\end{table}

In our original experiments, we fixed the pretrained label embeddings over the phase of training the universal model. Here, we present a method to jointly train the segmentation model as well as the label embeddings. We initialize the embeddings with the values computed from the pretrained networks, and make use of the following exponentially weighted average rule to update the centroids at the $t^{th}$ time step. 
\begin{equation}
    \mathrm{c}^{(k)}_{t} = \theta \cdot \mathrm{c}^{(k)}_{t-1} + (1-\theta) \cdot \mu_L(\mathcal{F}_t(x_u^{(k)})).
    \label{eq:centroid_update}
\end{equation}
In~\eqref{eq:centroid_update}, $\mathrm{c}^{(k)}_{t-1}$ denotes the centroids at the $(t-1)^{\text{th}}$ time step, $\mathcal{F}_t$ is the state of the encoder module at the $t^{th}$ time step and $\mu_L$ calculates the class wise centroids. $\theta$ is the update factor, where a value of $\theta=1$ implies that the centroids are not updated from their initial state, and a value of $\theta=0$ means that the centroids are calculated afresh at each update. We make an update to the centroids after every mini-batch of the original training data.

From~\tabref{tab:table}, experiments with $\theta=0.5$ suggests that jointly training the network as well as updating the label embeddings reduces the performance compared to having fixed label embeddings. We believe that this is primarily due to having insufficient training data for jointly updating embeddings as well as the network weights, although this merits a deeper investigation.


\begin{table*}[!t]
\centering

\resizebox{0.96\textwidth}{!}{
\begin{tabular}{@{} l *{19}{c} c c @{}}
    \toprule  \\[-1em]
    Method & 
    \rotatebox{90}{\footnotesize Road} & \rotatebox{90}{\footnotesize SideWalk} & \rotatebox{90}{\footnotesize Building} & \rotatebox{90}{\footnotesize Wall} & \rotatebox{90}{\footnotesize Fence} & \rotatebox{90}{\footnotesize Pole} & \rotatebox{90}{\footnotesize Traffic light} & \rotatebox{90}{\footnotesize Traffic Sign} & \rotatebox{90}{\footnotesize Vegetation} & \rotatebox{90}{\footnotesize Train} & \rotatebox{90}{\footnotesize Sky} & \rotatebox{90}{\footnotesize Person} & \rotatebox{90}{\footnotesize Rider} & \rotatebox{90}{\footnotesize Car} & \rotatebox{90}{\footnotesize Truck} & \rotatebox{90}{\footnotesize Bus} & \rotatebox{90}{\footnotesize Train} & \rotatebox{90}{\footnotesize MotorCycle} & \rotatebox{90}{\footnotesize Bicycle} && 
    \rotatebox{90}{ mIoU} \\ 
    \midrule \\[-1.5ex]
    K=1 & 
    92.18 & 51.29 & 80.07 & 00.00 & \textbf{24.01} & \textbf{33.73} & \textbf{26.16} & \textbf{38.71} & 82.30 & 36.39 & \textbf{81.61} & 54.38 & 20.48 & \textbf{81.71} & 02.37 & \textbf{22.79} & 03.85 & 01.31 & 46.23 
    && 41.03 \\
    
    K=3 & 
    \textbf{93.10} & \textbf{56.82} & 80.48 & 00.03 & 17.84 & 32.49 & 24.07 & 33.51 & \textbf{82.52} & \textbf{38.52} & 80.12 & 53.22 & 15.35 & 81.34 & \textbf{07.79} & 20.79 & 04.18 & \textbf{22.57} & \textbf{49.90 }
    && \textbf{41.82} \\
   
    K=5 & 
    89.58 & 48.50 & \textbf{81.21} & \textbf{14.55} & 07.89 & 27.77 & 22.72 & 33.95 & 84.36 & 34.33 & 80.52 & \textbf{54.39} & \textbf{22.51} & 81.52 & 02.41 & 09.43 & \textbf{07.28} & 10.40 & 48.18 
    && 40.08 \\
    
   \bottomrule
\end{tabular}}

\vspace{1em}
\small
\resizebox{0.96\textwidth}{!}{
\begin{tabular}{@{} l*{11}{c} c c @{}}
    \toprule  \\[-1em]
    Method & 
    \rotatebox{90}{\footnotesize Sky} & \rotatebox{90}{\footnotesize Buil.} & \rotatebox{90}{\footnotesize Pole} & \rotatebox{90}{\footnotesize Road} & \rotatebox{90}{\footnotesize Pave.} & \rotatebox{90}{\footnotesize Tree} & \rotatebox{90}{\footnotesize Sign} & \rotatebox{90}{\footnotesize Fence} & \rotatebox{90}{\footnotesize Car} & \rotatebox{90}{\footnotesize Ped.} & \rotatebox{90}{\footnotesize Bicy.} && \rotatebox{90}{\footnotesize mIoU} \\ 
    \midrule \\[-1.5ex]
    
    K=1 & 
     86.3 &  77.23 &  \textbf{17.13} &  84.99 &  53.35 &  70.57 &  31.99 &  32.45 &  72.94 &  36.61 &  \textbf{37.22}
     && 54.61 \\
    
    K=3 &
    \textbf{87.67} &  \textbf{78.51 }&  16.37 &  84.84 &  53.18 &  \textbf{73.33} &  \textbf{34.02} &  27.71 &  74.42 &  \textbf{40.36} &  34.24
     && 54.96 \\
    
    K=5 & 
    86.07 &  76.39 &  15.25 & \textbf{ 87.87} &  \textbf{63.6} &   70.95 &  32.6 &   \textbf{34.6} &   \textbf{76.63} &  30.42 &  30.9
     && \textbf{55.02} \\
    
    \bottomrule \\[-1.5ex]
    
\end{tabular}}
\vspace{-1em}
\captionsetup{width=0.95\textwidth}
\caption{Class-wise IoU values for the 19 classes in Cityscapes dataset and 11 classes in the CamVid dataset for different K, for N=100 on Resnet-18.}
\label{tab:class_wise_iou_supp}
\vspace{-1em}
\end{table*}

\subsection{Multiple Label Embeddings}

Many labels in a segmentation dataset often appear in more than one visual form or modalities. For example, \textit{road} class can appear as dry road, wet road, shady road etc., or a class labeled as \textit{building} can come in different structures and sizes. To better capture the multiple modalities involved in the visual information of the label, we propose using multiple embeddings for each label instead of a single mean centroid. This is analogous to polysemy in vocabulary, where many words can have multiple meanings and can occur in different contexts, and context specific word vector representations are used to capture this behavior. To calculate the multiple label embeddings, we perform K-means clustering of the pixel level encoder feature representations calculated from networks pretrained on the limited supervised data, and calculate similarity scores with all the multiple label embeddings.

\tabref{tab:table} shows that using K=3 embeddings per label gives an advantage over using 1 embedding per label, so apparently some amount of over segmentation helps. However, further increasing K to 5 hurts the performance, as not all the labels benefit from having multiple modalities per label. So, an interesting future direction can be to examine optimum number of embeddings per label. 

Particularly, from~\tabref{tab:class_wise_iou_supp}, it is evident that classes like \textit{Road}, \textit{Building}, \textit{Person} etc. benefit largely from having multiple embeddings per label.

\subsection{Choice of label embeddings}

In our work, we chose the pixel level class prototypes to be the label embeddings. We believe that this helps in better capturing visual information from the images compared to other approaches like Word2Vec~\cite{mikolov2013distributed}. To this end, we provide results of our approach replacing the prototype label embeddings with word vectors of the labels, by using the publicly available 128 dimensional word vectors for the labels from the Cityscapes and CamVid datasets.

From \tabref{tab:table}, having class prototypes as label embeddings, which are computed from the labeled data, performs better than using Word2vec based embeddings, which capture semantics of the word meaning rather than the visual appearance of the label. The performance improvement is more evident in case of N=100, which demonstrates that in presence of sufficient labeled data, class prototypes are better suited as label embeddings than word vector representations. Similar observations have been made in~\cite{snell2017prototypical} as well.

\section{Direct Softmax Entropy Regularization}

Entropy regularization is used to enhance the confidence of predictions made on unlabeled samples.
In the case of deep neural networks, applying this directly to the softmax scores will make the predictions confident by simply increasing the weights of the last layer. So, we follow an approach where we calculate similarity between normalized label prototypes and encoder embeddings through our entropy module. \textit{Direct SER} result from Table~\ref{tab:table} further demonstrates the fact that applying SER (softmax entropy regularization) directly to our network shows inferior performance compared to the proposed entropy module based approach.



\end{document}